\documentclass[11pt,letterpaper]{article}
\PassOptionsToPackage{table}{xcolor}
\usepackage[T1]{fontenc}
\usepackage{tgheros,tgcursor}
\usepackage{newpxtext,newpxmath}

\usepackage[letterpaper,left=1.05in,right=1.05in,top=.85in,bottom=.85in,
  headheight=14pt,headsep=16pt]{geometry}
\usepackage{microtype}
\usepackage[hyphens]{url}
\usepackage{xurl}
\usepackage{graphicx}
\usepackage{amsmath}
\usepackage{booktabs,array}
\usepackage[table]{xcolor}
\usepackage{natbib}
\usepackage{caption}
\usepackage{tikz}
\usepackage{float}
\usepackage{needspace}
\usepackage{titlesec}
\usepackage{fancyhdr}
\usepackage{etoolbox}
\usepackage{hyperref}
\definecolor{TrecInk}{HTML}{17242B}
\definecolor{TrecMuted}{HTML}{52616B}
\definecolor{TrecRule}{HTML}{D5DFE1}
\hypersetup{colorlinks=true,linkcolor=TrecInk,citecolor=TrecTeal,
  urlcolor=TrecTeal,bookmarksnumbered=true,
  pdftitle={TReC: Comparing Move-Induced Representation Changes in Go},
  pdfauthor={Zhuangsheng Li, Yunhan Zhang, Dailin Li, Qingxian Yu, Yu Wu, Youqi Song, Bangyuan He}}
\titleformat{\section}{\large\bfseries\color{TrecInk}}
  {\color{TrecTeal}\thesection}{.65em}{}
\titleformat{\subsection}{\normalsize\bfseries\color{TrecInk}}
  {\thesubsection}{.65em}{}
\titleformat{\paragraph}[runin]{\normalsize\bfseries}{}{0pt}{}
\titlespacing*{\section}{0pt}{16pt plus 3pt minus 2pt}{7pt}
\titlespacing*{\subsection}{0pt}{12pt plus 2pt minus 1pt}{5pt}
\titlespacing*{\paragraph}{0pt}{7pt plus 1pt}{.6em}
\AtBeginEnvironment{tabular*}{}
\let\TrecOriginalTable\table
\renewcommand{\table}[1][htbp]{\TrecOriginalTable[H]}

\fancypagestyle{plain}{\fancyhf{}\fancyfoot[C]{\small\color{TrecMuted}\thepage}
  }
\input{preprint/frontmatter}
\definecolor{TrecTeal}{HTML}{216C70}
\definecolor{TrecLight}{HTML}{EFF5F5}
\usetikzlibrary{arrows.meta,positioning}
\newcommand{\TrecAbsGraphic}[1]{%
  \includegraphics[#1]{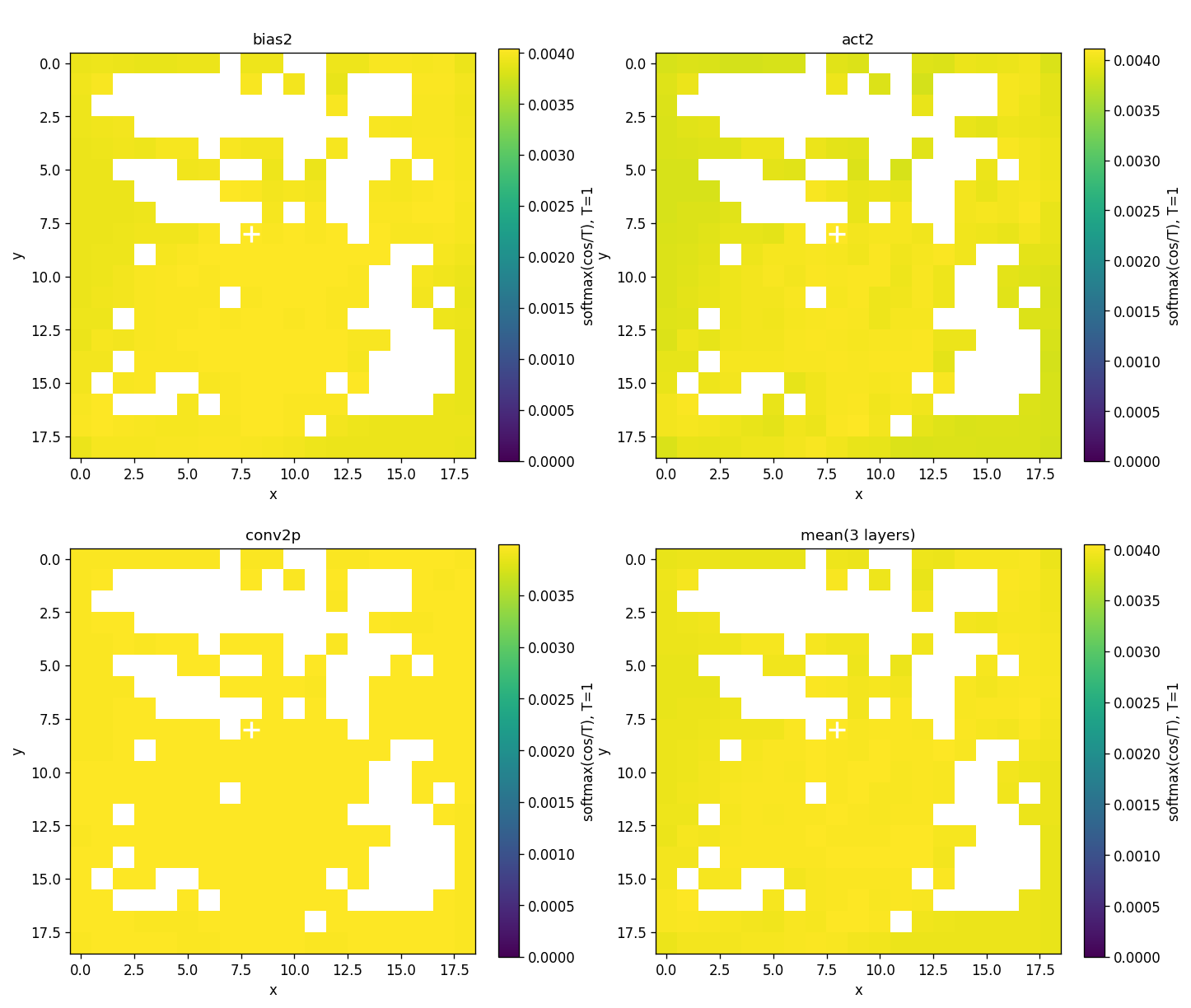}}
\newcommand{\TrecDeltaGraphic}[1]{%
  \includegraphics[#1]{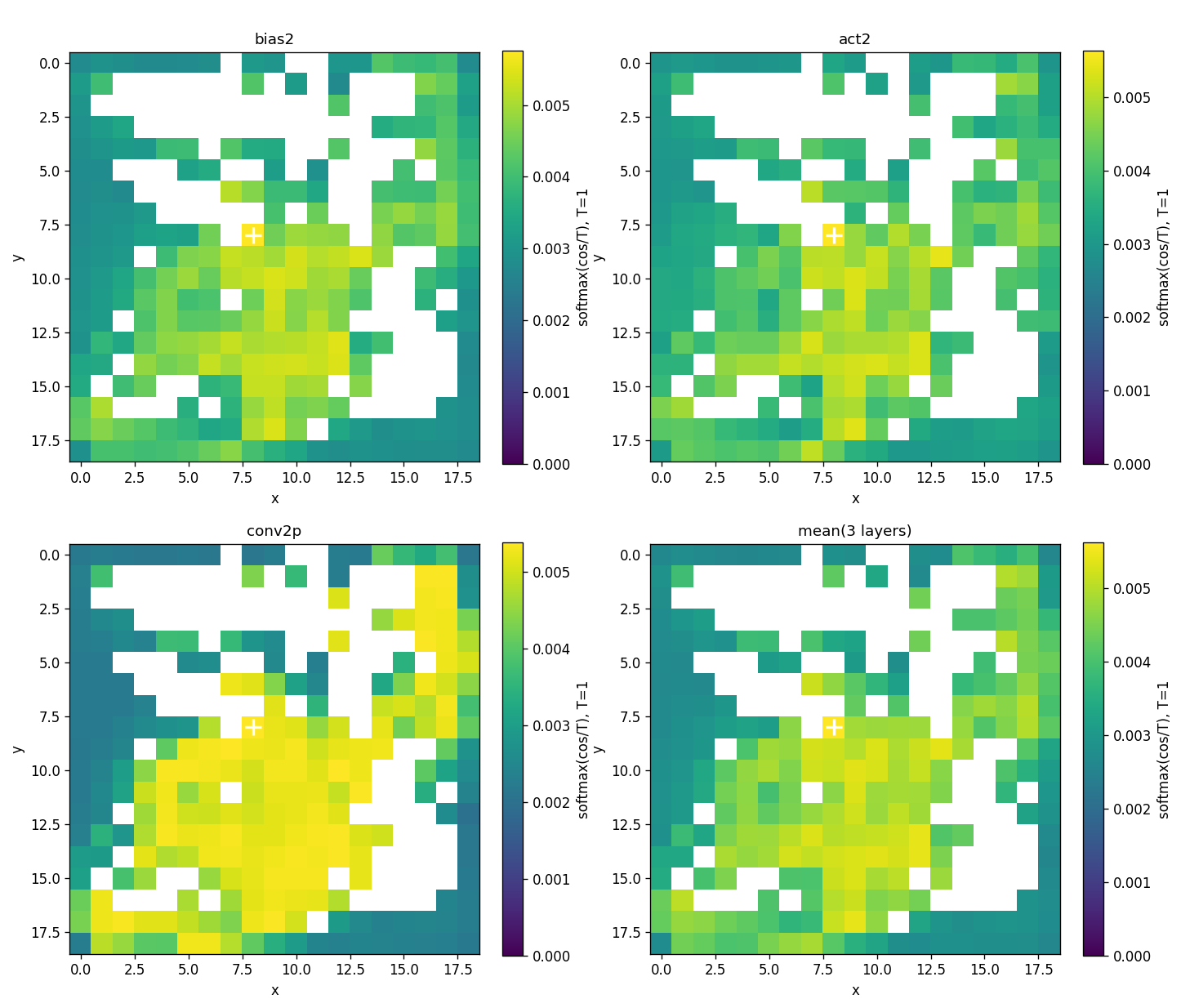}}
\newcommand{\TrecBoard}[3]{%
  #1{width=#3,trim=#2,clip}}
\newcommand{\TrecCandidateMarks}[1]{%
  \pgfmathsetmacro{\TrecU}{#1/19}%
  \foreach \TrecR/\TrecC in {9/8,8/9}{%
    \draw[black,line width=1.2pt] ({\TrecC*\TrecU},{#1-\TrecR*\TrecU}) rectangle ++(\TrecU,-\TrecU);
    \draw[white,line width=.5pt] ({\TrecC*\TrecU},{#1-\TrecR*\TrecU}) rectangle ++(\TrecU,-\TrecU);}%
  \draw[black,line width=1.2pt] ({13.5*\TrecU},{#1-9.5*\TrecU}) circle[radius=.9*\TrecU];
  \draw[white,line width=.5pt] ({13.5*\TrecU},{#1-9.5*\TrecU}) circle[radius=.9*\TrecU];}

\newcommand{\TrecOverviewFigure}{%
\begin{figure}[H]
\centering
\begin{tikzpicture}[
  font=\small,
  state/.style={draw=TrecRule,rounded corners=3pt,fill=white,
    minimum width=23mm,minimum height=9mm,inner sep=4pt},
  change/.style={draw=TrecTeal!65,rounded corners=3pt,fill=TrecLight,
    minimum width=27mm,minimum height=9mm,inner sep=4pt},
  flow/.style={-{Stealth[length=5pt]},draw=TrecMuted,line width=.65pt}
]
\node[state,minimum width=18mm,align=center] (initial) at (0,0) {Shared\\state $s$};
\node[state] (expert) at (3.1,.8) {$T(s,a^\star)$};
\node[state] (predicted) at (3.1,-.8) {$T(s,\hat a)$};
\node[change] (dexpert) at (6.6,.8) {$\Delta_l(s,a^\star)$};
\node[change] (dpredicted) at (6.6,-.8) {$\Delta_l(s,\hat a)$};
\node[change,minimum width=23mm,align=center] (score) at (10.0,0) {Cosine\\\textbf{TReC}};
\node[align=center,font=\sffamily\small\color{TrecMuted}] at (3.1,1.85) {Execute Go rules};
\node[align=center,font=\sffamily\small\color{TrecMuted}] at (6.6,1.85) {Encode with $h_l$\\subtract $h_l(s)$};
\draw[flow] (initial.east) -- node[pos=.47,above,inner sep=3pt] {$a^\star$} (expert.west);
\draw[flow] (initial.east) -- node[pos=.47,below,inner sep=3pt] {$\hat a$} (predicted.west);
\draw[flow] (expert.east) -- (dexpert.west);
\draw[flow] (predicted.east) -- (dpredicted.west);
\draw[flow] (dexpert.east) -- (score.west);
\draw[flow] (dpredicted.east) -- (score.west);
\end{tikzpicture}
\caption{Two legal moves produce rule-executed successors of the same
state. TReC compares their changes in a shared board-state representation
$h_l$, relative to the shared initial representation.}
\label{fig:overview}
\end{figure}}
\newcommand{\TrecGalleryMap}[4]{%
\begin{tikzpicture}[x=1pt,y=1pt,font=\footnotesize]
\node[anchor=south west,inner sep=0pt] at (0,0)
  {\TrecBoard{#1}{#2}{80pt}};
\node[anchor=south west,inner sep=0pt] at (85,0)
  {#1{width=3pt,height=80pt,trim=#3,clip}};
\node[anchor=north west,inner sep=0pt,text=TrecMuted] at (93,80) {High};
\node[anchor=south west,inner sep=0pt,text=TrecMuted] at (93,0) {Low};
\node[anchor=base,inner sep=0pt] at (40,87) {\small #4};
\end{tikzpicture}}
\newcommand{\TrecLayerGallery}{%
\begin{figure}[H]
\centering
\setlength{\tabcolsep}{0pt}
\begin{tabular*}{\linewidth}{@{\extracolsep{\fill}}cccc@{}}
\multicolumn{4}{@{}l@{}}{\sffamily\small\bfseries (a) Successor states}\\[5pt]
\TrecGalleryMap{\TrecAbsGraphic}{51.0bp 400.8bp 521.4bp 38.4bp}{364.8bp 400.8bp 490.8bp 38.4bp}{\texttt{bias2}} &
\TrecGalleryMap{\TrecAbsGraphic}{477.6bp 400.8bp 94.8bp 38.4bp}{791.4bp 400.8bp 64.2bp 38.4bp}{\texttt{act2}} &
\TrecGalleryMap{\TrecAbsGraphic}{51.0bp 39.6bp 521.4bp 399.6bp}{364.8bp 39.6bp 490.8bp 399.6bp}{\texttt{conv2p}} &
\TrecGalleryMap{\TrecAbsGraphic}{477.6bp 39.6bp 94.8bp 399.6bp}{791.4bp 39.6bp 64.2bp 399.6bp}{Layer mean}\\[10pt]
\multicolumn{4}{@{}l@{}}{\sffamily\small\bfseries (b) State changes}\\[5pt]
\TrecGalleryMap{\TrecDeltaGraphic}{51.0bp 401.4bp 515.4bp 38.4bp}{370.8bp 401.4bp 484.8bp 38.4bp}{\texttt{bias2}} &
\TrecGalleryMap{\TrecDeltaGraphic}{477.6bp 401.4bp 88.8bp 38.4bp}{797.4bp 401.4bp 58.2bp 38.4bp}{\texttt{act2}} &
\TrecGalleryMap{\TrecDeltaGraphic}{51.0bp 39.6bp 515.4bp 399.6bp}{370.8bp 39.6bp 484.8bp 399.6bp}{\texttt{conv2p}} &
\TrecGalleryMap{\TrecDeltaGraphic}{477.6bp 39.6bp 88.8bp 399.6bp}{797.4bp 39.6bp 58.2bp 399.6bp}{Layer mean}
\end{tabular*}
\caption{All representation panels from the original heatmaps,
arranged by representation: \textbf{(a)} direct successor-state similarity
and \textbf{(b)} transition representation similarity.
Each panel retains its original colors and independently scaled color bar;
High and Low denote that panel's local range.
Colors encode softmax-normalized cosine scores at temperature $1$.
White locations are masked and the cross marks the reference action.
The complete original plots with their numerical axes and scales remain
in the source figure files.}
\label{fig:layers}
\end{figure}}
\newcommand{\TrecEvidenceMap}[3]{%
\begin{tikzpicture}[x=1pt,y=1pt,font=\footnotesize]
\node[anchor=south west,inner sep=0pt] at (0,0)
  {\TrecBoard{#1}{#2}{112pt}};
\TrecCandidateMarks{112}
\node[anchor=south west,inner sep=0pt] at (0,-8)
  {#1{width=3pt,height=112pt,angle=-90,trim=#3,clip}};
\node[anchor=north west,inner sep=0pt,text=TrecMuted] at (0,-12) {Low};
\node[anchor=north east,inner sep=0pt,text=TrecMuted] at (112,-12) {High};
\end{tikzpicture}}
\newcommand{\TrecEvidenceFigure}{%
\begin{figure}[!htbp]
\centering
\begin{minipage}[t]{.33\linewidth}
\centering
{\sffamily\small\bfseries (a) Successor states\par}\vspace{7pt}
\TrecEvidenceMap{\TrecAbsGraphic}{477.6bp 39.6bp 94.8bp 399.6bp}{791.4bp 39.6bp 64.2bp 399.6bp}
\end{minipage}\hspace{.08\linewidth}%
\begin{minipage}[t]{.33\linewidth}
\centering
{\sffamily\small\bfseries (b) State changes\par}\vspace{7pt}
\TrecEvidenceMap{\TrecDeltaGraphic}{477.6bp 39.6bp 88.8bp 399.6bp}{797.4bp 39.6bp 58.2bp 399.6bp}
\end{minipage}
\caption{Two cosine scores over the unmasked intersections of one position
(softmax-normalized, $T=1$; mean of three evaluator layers).
Each panel has its own color scale, so colors show ranking within a
panel. White locations are masked; the cross marks the reference.
Squares mark the two unmasked intersections adjacent to the
reference and the circle a candidate at distance $5.1$
(Table~\ref{tab:cells}).}
\label{fig:evidence}
\end{figure}}

\title{TReC: Comparing Move-Induced\\Representation Changes in Go}

\author{
    Zhuangsheng Li\textsuperscript{\rm 1}\equalcontrib,
    Yunhan Zhang\textsuperscript{\rm 2}\equalcontrib,
    Dailin Li\textsuperscript{\rm 3}\equalcontrib,\\
    Qingxian Yu\textsuperscript{\rm 4}\equalcontrib,
    Yu Wu\textsuperscript{\rm 1}\equalcontrib\thanks{Corresponding author.},
    Youqi Song\textsuperscript{\rm 5},
    Bangyuan He\textsuperscript{\rm 6}
}
\affiliations{
    \textsuperscript{\rm 1}University of California, San Diego\quad
    \textsuperscript{\rm 2}University of Melbourne\\
    \textsuperscript{\rm 3}University of Virginia\quad
    \textsuperscript{\rm 4}Shanghai Jiao Tong University\\
    \textsuperscript{\rm 5}East China Normal University\quad
    \textsuperscript{\rm 6}New York University
}

\begin{document}
\maketitle
\begin{abstract}
When a Go model predicts a move different from the expert's, coordinate
distance measures how far the prediction lands from the reference.
We introduce Transition Representation Consistency (TReC) to compare
how the two moves change a board-state embedding.
Both moves are executed from the same position under the game rules,
and TReC is the cosine similarity between the two resulting
representation changes.
In experiments with a frozen KataGo network on 680 mid-to-late-game
positions from five games, the gap between
the highest and lowest candidate per position has a median of
$0.01$ to $0.08$ for the direct successor-state cosine and $0.88$ to
$1.33$ for TReC, depending on the evaluator layer.
TReC follows coordinate distance only weakly: the rank correlation
between a candidate's distance to the reference and its TReC, computed
within each position, has a median of $-0.19$ over these positions.
\end{abstract}

\section{Introduction}
When a language model plays Go, it predicts a move from a board state,
and the prediction is scored against a reference.
Work on Go move prediction reports how often the model reproduces the
move an expert played \citep{clark2015training}.
LoGos relaxes the single reference: KataGo-Bench-1K counts a prediction
as correct when it falls among several candidate moves annotated with
KataGo, and the training reward uses KataGo's ranking and win-rate
estimates \citep{ma2025logos}.
A candidate-set hit says whether the prediction is an accepted move; a
win-rate estimate says how strong it is.
When a prediction differs from the expert move, one more thing is worth
knowing besides how far away it landed and how strong it is: whether it
changes the position in a similar way.
Coordinate distance cannot answer this.
A move that fills one of a group's last two liberties puts the group in
atari, while a move on a neighboring intersection may not touch the
group.
We therefore ask: \emph{how similar are the representation changes that
two moves cause when played from the same board state?}

Go rules determine legal actions and their successor states; a
board-state embedding model supplies their representations.
Two successors of the same position can retain much of the same board
structure, so in some positions their direct cosine similarities may
concentrate in a narrow band.
Transition Representation Consistency (TReC) instead compares the
directions of two successor-minus-initial representation vectors.
The symbolic component executes the proposed and reference actions;
the neural component compares their encoded changes.
Rule execution is exact, and only the similarity depends on the
representation.

This short paper defines TReC, analyzes what the subtraction removes,
and examines the candidate moves of 680 positions using KataGo
\citep{wu2019katago}, where TReC follows coordinate distance only weakly.

\section{Related Work}
\paragraph{Language-conditioned play.}
Language-conditioned agents \citep{hu2023instructrl} and ChessGPT
\citep{feng2023chessgpt} connect language with action selection.
In Go, LoGos combines expert knowledge with language-model reasoning
and is evaluated against multiple KataGo-annotated candidate moves
\citep{ma2025logos}.
\citet{chen2026atari} build controlled atari and sente tasks and find
that a fine-tuned LoGos struggles to recognize changes in the atari
relationship.
These tasks test named concepts; TReC compares two moves without
naming one.

\paragraph{Comparing actions through their effects.}
FeUdal Networks reward a worker with the cosine between a latent state
displacement and a goal direction set by a manager
\citep{vezhnevets2017feudal}.
\citet{baram2021action} define action redundancy through the next
states that actions induce; under deterministic dynamics, two actions
are redundant when they lead to the same state.
\citet{chandak2019action} learn action embeddings from observed
transitions so that a policy can generalize across actions with similar
effects.
In video understanding, \citet{parmar2025cate} select the action whose
representation best matches a state-change representation by cosine
similarity.
Bisimulation metrics \citep{ferns2004metrics} and the representations
trained with them \citep{zhang2021bisimulation} organize states by
behavioral similarity.
The vector difference itself follows representation arithmetic
\citep{mikolov-etal-2013-linguistic}.
TReC takes the cosine of two displacements from one position, for
evaluation, using the same encoder for both moves.

\paragraph{Representations of game-playing networks.}
Probing studies ask what a game network encodes about a position:
human chess concepts in AlphaZero \citep{mcgrath2022acquisition} and
concepts from natural-language commentary in Go-playing networks
\citep{tomlin2022understanding}.
Similarity indices such as CKA compare the representations of two
networks over a set of inputs \citep{kornblith2019similarity}.
TReC stays within one network and compares the changes that two moves
cause from the same position.

\section{Transition Representation Consistency}
\label{sec:trec}
Let $s$ be a Go state with the information required by the selected
rules, $\hat a$ a predicted move, and $a^\star$ the recorded expert
move. For a legal move $a$, let $T(s,a)$ be its rule-defined
successor. A shared board-state embedding model provides a vector
$h_l(s)$ at layer $l$.

\TrecOverviewFigure

\paragraph{Definition.}
Figure~\ref{fig:overview} outlines the comparison. Define
\begin{equation}
\Delta_l(s,a)=h_l\!\left(T(s,a)\right)-h_l(s).
\label{eq:delta}
\end{equation}
The layer-wise score is
\begin{equation}
\operatorname{TReC}_l(s,\hat a,a^\star)
=
\frac{\Delta_l(s,\hat a)^\top\Delta_l(s,a^\star)}
{\|\Delta_l(s,\hat a)\|_2\,\|\Delta_l(s,a^\star)\|_2}.
\label{eq:trec}
\end{equation}
Equation~\ref{eq:trec} is defined when both differences are nonzero.
Its range is $[-1,1]$; normalization makes the score depend on
change direction rather than magnitude. It also applies to any legal candidate $a$.
Combining layers additionally requires a layer set and an aggregation
rule.

For nonzero successor representations, direct state similarity is
\begin{equation}
S_l^{\mathrm{abs}}(s,a,a^\star)
=
\cos\!\left(h_l(T(s,a)),h_l(T(s,a^\star))\right).
\label{eq:absolute}
\end{equation}
\paragraph{Why subtract the initial state?}
For any constant vector $c$, replacing every representation by
$h'_l(x)=h_l(x)+c$ leaves $\Delta_l$ and TReC unchanged.
Direct successor cosine does not have this invariance.
For fixed successor vectors $x,y$ and any nonzero $c$,
\[
\lim_{t\to\infty}\cos(x+tc,y+tc)=1,
\]
since both normalized vectors converge to $c/\|c\|_2$.
Thus, direct state similarity can approach one under a shared offset
while transition similarity stays fixed.
A common component that raises cosine similarity between unrelated
items is also documented for word and contextual embeddings
\citep{mu2018allbutthetop,ethayarajh2019contextual}.

\paragraph{What the subtraction does not remove.}
The subtraction cancels what both successors share with the initial
state. It does not cancel changes that every move causes, such as the
switch of the player to move or the shift of the move history.
If this shared change has size $M$ and the move-specific parts are
orthogonal unit vectors, then $\Delta_l(s,a)=(M,1,0)$ and
$\Delta_l(s,a')=(M,0,1)$ have cosine $M^2/(M^2+1)$, which approaches
one as $M$ grows.
TReC values are therefore comparable across candidates at one position
and layer; their absolute level depends on how strongly the layer
encodes turn and history.

\paragraph{Interpretation.}
TReC is conditional on the reference action and the evaluator.
A higher score means the candidate changes the representation in a
direction closer to that of the reference.
Our experiments use a frozen KataGo policy network.

\section{Data}
The positions come from five professional games played in 2025--2026,
selected from GoGoD \citep{gogodonline}.
SGF parsing reconstructs positions, history, and rule information.
We use every position from move 51 on that has a following move; the
move played next is its reference $a^\star$.
For intersection moves, the distance between a candidate and the
reference is
$d(a,a^\star)=\|p(a)-p(a^\star)\|_2$,
where $p(a)$ is the board coordinate, so it is measured in
intersections.

\section{Results}
We look at how the score varies over the candidate moves of a
position.
Figure~\ref{fig:evidence} shows both similarities over the 251 unmasked
intersections of one position, and Table~\ref{tab:cells} gives the
cosines of the marked candidates.
Colors are softmax-normalized separately per panel, so they show
ranking within a panel, not magnitude.

\TrecEvidenceFigure

\begin{table}[tb]
\centering
\small
\setlength{\tabcolsep}{2.5pt}
\begin{tabular*}{\linewidth}{@{\extracolsep{\fill}}lrrrrrr@{}}
\toprule
 & \multicolumn{3}{c}{TReC} & \multicolumn{3}{c}{Successor cosine}\\
\cmidrule(lr){2-4}\cmidrule(l){5-7}
Candidate & \texttt{bias2} & \texttt{act2} & \texttt{conv2p} & \texttt{bias2} & \texttt{act2} & \texttt{conv2p}\\
\midrule
$\circ$ $d=5.1$ & $0.943$ & $0.962$ & $0.973$ & $0.999$ & $0.996$ & $1.000$\\
$\square$ below & $0.891$ & $0.895$ & $0.972$ & $0.998$ & $0.986$ & $0.999$\\
$\square$ right & $0.755$ & $0.831$ & $0.966$ & $0.992$ & $0.983$ & $0.999$\\
\midrule
Minimum & $0.248$ & $0.328$ & $0.018$ & $0.967$ & $0.933$ & $0.993$\\
Median & $0.639$ & $0.649$ & $0.817$ & $0.989$ & $0.971$ & $0.998$\\
\bottomrule
\end{tabular*}
\caption{Cosines with the reference move in three evaluator layers
for the position of Figure~\ref{fig:evidence}.
$\circ$: the candidate at distance $d=5.1$; $\square$: the two adjacent
candidates, by their side of the reference.
Minimum and median are over the 250 candidates other than the
reference.}
\label{tab:cells}
\end{table}

Two observations follow.
First, direct successor cosines lie in a narrow band and TReC covers a
wide range.
In this position the successor cosine of every candidate is above
$0.93$ in every layer, while TReC runs from one at the reference down
to between $0.02$ and $0.33$.
Over 680 mid-to-late-game positions (move 51 on) of the five games,
we take the difference between the highest and the lowest
candidate within each position.
Its median over the positions is $0.01$ to $0.08$ for the successor
cosine and $0.88$ to $1.33$ for TReC, depending on the layer.
The offset argument of Section~\ref{sec:trec} shows that a shared
component can produce such a band; these cosines cannot tell whether
one does here.
Second, TReC follows distance to the reference only weakly.
In the plotted position, the candidate at distance $5.1$ scores $0.94$
in \texttt{bias2} and $0.96$ in \texttt{act2}, above both adjacent
candidates ($0.76$ to $0.89$); in \texttt{conv2p} the three differ by
less than $0.01$.
For each of the 680 positions we compute the rank correlation between
the distance of a candidate to the reference and its TReC (three-layer
mean); the median of this correlation over the positions is $-0.19$.
The candidate with the highest TReC other than the reference lies two
or more intersections from it in $68\%$ of the positions.
For scale, a legal move drawn uniformly at random has a mean TReC of
$0.43$ over these positions (three-layer mean).

\section{Conclusion}
TReC combines rule execution with a neural representation to compare
what two moves change.
Across 680 positions from five games, the median within-position rank
correlation between TReC and distance to the reference move is $-0.19$.
The evidence is limited to these games and one evaluator network.
TReC is relative to one reference move and one evaluator, and it does
not measure move quality.

\bibliography{preprint/references}

@article{wu2019katago,
  author = {Wu, David J.},
  title = {Accelerating Self-Play Learning in {Go}},
  journal = {arXiv preprint arXiv:1902.10565},
  year = {2019}
}

@inproceedings{hu2023instructrl,
  author = {Hu, Hengyuan and Sadigh, Dorsa},
  title = {Language Instructed Reinforcement Learning for Human-{AI} Coordination},
  booktitle = {Proceedings of the 40th International Conference on Machine Learning},
  volume = {202},
  series = {Proceedings of Machine Learning Research},
  pages = {13584--13598},
  publisher = {PMLR},
  year = {2023}
}

@inproceedings{feng2023chessgpt,
  author = {Feng, Xidong and Luo, Yicheng and Wang, Ziyan and Tang, Hongrui and Yang, Mengyue and Shao, Kun and Mguni, David and Du, Yali and Wang, Jun},
  title = {{ChessGPT}: Bridging Policy Learning and Language Modeling},
  booktitle = {Advances in Neural Information Processing Systems},
  volume = {36},
  year = {2023}
}

@inproceedings{mikolov-etal-2013-linguistic,
  author = {Mikolov, Tomas and Yih, Wen-tau and Zweig, Geoffrey},
  title = {Linguistic Regularities in Continuous Space Word Representations},
  booktitle = {Proceedings of NAACL-HLT},
  pages = {746--751},
  publisher = {Association for Computational Linguistics},
  year = {2013}
}

@misc{gogodonline,
  author = {{GoGoD}},
  title = {{GoGoD Online: Games of Go on Download}},
  year = {n.d.},
  howpublished = {\url{https://gogodonline.co.uk/}},
  note = {Accessed November 5, 2026}
}

@inproceedings{ma2025logos,
  author = {Ma, Yichuan and Li, Linyang and Chen, Yongkang and Li, Peiji and Ye, Jiasheng and Guo, Qipeng and Lin, Dahua and Chen, Kai},
  title = {Mixing Expert Knowledge: Bring Human Thoughts Back To the Game of {Go}},
  booktitle = {Advances in Neural Information Processing Systems},
  volume = {38},
  year = {2025}
}

@article{chen2026atari,
  author = {Chen, Taiyi and Deng, Iskar and Paullada, Amandalynne},
  title = {Recognizing Atari and Assessing Sente: Controlled Evaluation Tasks for Language-Based {Go} Agents},
  journal = {arXiv preprint arXiv:2610.22953},
  year = {2026}
}

@inproceedings{zhang2021bisimulation,
  author = {Zhang, Amy and McAllister, Rowan and Calandra, Roberto and Gal, Yarin and Levine, Sergey},
  title = {Learning Invariant Representations for Reinforcement Learning without Reconstruction},
  booktitle = {International Conference on Learning Representations},
  year = {2021}
}

@inproceedings{vezhnevets2017feudal,
  author = {Vezhnevets, Alexander Sasha and Osindero, Simon and Schaul, Tom and Heess, Nicolas and Jaderberg, Max and Silver, David and Kavukcuoglu, Koray},
  title = {{FeUdal} Networks for Hierarchical Reinforcement Learning},
  booktitle = {Proceedings of the 34th International Conference on Machine Learning},
  series = {Proceedings of Machine Learning Research},
  volume = {70},
  pages = {3540--3549},
  publisher = {PMLR},
  year = {2017}
}

@inproceedings{baram2021action,
  author = {Baram, Nir and Tennenholtz, Guy and Mannor, Shie},
  title = {Action Redundancy in Reinforcement Learning},
  booktitle = {Proceedings of the Thirty-Seventh Conference on Uncertainty in Artificial Intelligence},
  series = {Proceedings of Machine Learning Research},
  volume = {161},
  pages = {376--385},
  publisher = {PMLR},
  year = {2021}
}

@inproceedings{parmar2025cate,
  author = {Parmar, Paritosh and Peh, Eric and Fernando, Basura},
  title = {Learning to Visually Connect Actions and their Effects},
  booktitle = {IEEE/CVF Winter Conference on Applications of Computer Vision (WACV)},
  pages = {1477--1487},
  doi = {10.1109/WACV61041.2025.00151},
  year = {2025}
}

@inproceedings{chandak2019action,
  author = {Chandak, Yash and Theocharous, Georgios and Kostas, James and Jordan, Scott and Thomas, Philip S.},
  title = {Learning Action Representations for Reinforcement Learning},
  booktitle = {Proceedings of the 36th International Conference on Machine Learning},
  series = {Proceedings of Machine Learning Research},
  volume = {97},
  pages = {941--950},
  publisher = {PMLR},
  year = {2019}
}

@inproceedings{ferns2004metrics,
  author = {Ferns, Norm and Panangaden, Prakash and Precup, Doina},
  title = {Metrics for Finite {Markov} Decision Processes},
  booktitle = {Proceedings of the 20th Conference on Uncertainty in Artificial Intelligence},
  pages = {162--169},
  year = {2004}
}

@article{mcgrath2022acquisition,
  author = {McGrath, Thomas and Kapishnikov, Andrei and Toma{\v{s}}ev, Nenad and Pearce, Adam and Wattenberg, Martin and Hassabis, Demis and Kim, Been and Paquet, Ulrich and Kramnik, Vladimir},
  title = {Acquisition of Chess Knowledge in {AlphaZero}},
  journal = {Proceedings of the National Academy of Sciences},
  volume = {119},
  number = {47},
  pages = {e2206625119},
  year = {2022}
}

@inproceedings{tomlin2022understanding,
  author = {Tomlin, Nicholas and He, Andre and Klein, Dan},
  title = {Understanding Game-Playing Agents with Natural Language Annotations},
  booktitle = {Proceedings of the 60th Annual Meeting of the Association for Computational Linguistics (Volume 2: Short Papers)},
  pages = {797--807},
  year = {2022}
}

@inproceedings{kornblith2019similarity,
  author = {Kornblith, Simon and Norouzi, Mohammad and Lee, Honglak and Hinton, Geoffrey},
  title = {Similarity of Neural Network Representations Revisited},
  booktitle = {Proceedings of the 36th International Conference on Machine Learning},
  series = {Proceedings of Machine Learning Research},
  volume = {97},
  pages = {3519--3529},
  publisher = {PMLR},
  year = {2019}
}

@inproceedings{mu2018allbutthetop,
  author = {Mu, Jiaqi and Viswanath, Pramod},
  title = {All-but-the-Top: Simple and Effective Postprocessing for Word Representations},
  booktitle = {International Conference on Learning Representations},
  year = {2018}
}

@inproceedings{ethayarajh2019contextual,
  author = {Ethayarajh, Kawin},
  title = {How Contextual are Contextualized Word Representations? {Comparing} the Geometry of {BERT}, {ELMo}, and {GPT-2} Embeddings},
  booktitle = {Proceedings of the 2019 Conference on Empirical Methods in Natural Language Processing and the 9th International Joint Conference on Natural Language Processing},
  pages = {55--65},
  year = {2019}
}

@inproceedings{clark2015training,
  author = {Clark, Christopher and Storkey, Amos},
  title = {Training Deep Convolutional Neural Networks to Play {G}o},
  booktitle = {Proceedings of the 32nd International Conference on Machine Learning},
  series = {Proceedings of Machine Learning Research},
  volume = {37},
  pages = {1766--1774},
  year = {2015}
}

\appendix
\setcounter{table}{0}
\setcounter{figure}{0}
\renewcommand{\thetable}{S\arabic{table}}
\renewcommand{\thefigure}{S\arabic{figure}}
\section*{Appendix}
\section{Evaluation Record}
Figure~\ref{fig:evidence} and Figure~\ref{fig:layers} use three evaluator
representations, labeled \texttt{bias2}, \texttt{act2}, and
\texttt{conv2p} in the KataGo network, and their mean.
The candidate-level statistics cover every position of the five
games from move 51 on that has a following move, with komi $7.5$.
Of these 695 positions, 15 contain a ko, fail the check of the rule
replay against the stored record, and are skipped, which leaves 680.
For each of them, the game is replayed from its recorded moves under
the rules up to the position, and the position is encoded from this
replay.
The reference successor and every candidate successor are obtained by
playing the corresponding move on a copy of the replayed position and
encoding the result in the same way, so the reference move scores
exactly one.

\TrecLayerGallery

\end{document}